\documentclass[sn-mathphys,Numbered]{sn-jnl}% Math and Physical Sciences Reference Style
\usepackage[utf8]{inputenc}
\usepackage{graphicx}%
\usepackage{subfigure}
\usepackage{float}%
\usepackage{multirow}%
\usepackage{amsmath,amssymb,amsfonts}%
\usepackage{amsthm}%
\usepackage{mathrsfs}%
\usepackage[title]{appendix}%
\usepackage{xcolor}%
\usepackage{textcomp}%
\usepackage{manyfoot}%
\usepackage{booktabs}%
\usepackage{algorithm}%
\usepackage{algorithmicx}%
\usepackage{algpseudocode}%
\usepackage{listings}%
\usepackage{verbatim}
\usepackage{lscape}
\usepackage{pdflscape}

\begin{document}

\title[Article Title]{Mass Spectra Prediction with Structural Motif-based Graph Neural Networks}

%%=============================================================%%
%% Prefix	-> \pfx{Dr}
%% GivenName	-> \fnm{Joergen W.}
%% Particle	-> \spfx{van der} -> surname prefix
%% FamilyName	-> \sur{Ploeg}
%% Suffix	-> \sfx{IV}
%% NatureName	-> \tanm{Poet Laureate} -> Title after name
%% Degrees	-> \dgr{MSc, PhD}
%% \author*[1,2]{\pfx{Dr} \fnm{Joergen W.} \spfx{van der} \sur{Ploeg} \sfx{IV} \tanm{Poet Laureate} 
%%                 \dgr{MSc, PhD}}\email{iauthor@gmail.com}
%%=============================================================%%

\author[1,2]{\fnm{Jiwon} \sur{Park}}\email{jwpark21@snu.ac.kr}

\author*[3]{\fnm{Jeonghee} \sur{Jo}}\email{jh.jo@kist.re.kr}
%%\equalcont{These authors contributed equally to this work.}

\author*[1,4,5]{\fnm{Sungroh} \sur{Yoon}}\email{sryoon@snu.ac.kr}
%%\equalcont{These authors contributed equally to this work.}

\affil[1]{\orgdiv{Interdisciplinary Program in Artificial Intelligence}, \orgname{Seoul National University}, \orgaddress{\city{Seoul}, \postcode{08826}, \country{Republic of Korea}}}

\affil[2]{\orgname{LG Chem}, \orgaddress{\city{Seoul}, \postcode{07795}, \country{Republic of Korea}}}

\affil[3]{\orgdiv{Center for Neuromorphic Engineering}, \orgname{Korea Institute of Science and Technology (KIST)}, \orgaddress{\city{Seoul}, \postcode{02456}, \country{Republic of Korea}}}

\affil[4]{\orgdiv{Department of Electrical and Computer Engineering}, \orgname{Seoul National University}, \orgaddress{\city{Seoul}, \postcode{08826}, \country{Republic of Korea}}}

\affil[5]{\orgdiv{Artificial Intelligence Institute}, \orgname{Seoul National University}, \orgaddress{\city{Seoul}, \postcode{08826}, \country{Republic of Korea}}}

%%==================================%%
%% sample for unstructured abstract %%
%%==================================%%

\abstract{ 
Mass spectra, which are agglomerations of ionized fragments from targeted molecules, play a crucial role across various fields for the identification of molecular structures. A prevalent analysis method involves spectral library searches, where unknown spectra are cross-referenced with a database. The effectiveness of such search-based approaches, however, is restricted by the scope of the existing mass spectra database, underscoring the need to expand the database via mass spectra prediction. In this research, we propose the Motif-based Mass Spectrum Prediction Network (MoMS-Net), a system that predicts mass spectra using the information derived from structural motifs and the implementation of Graph Neural Networks (GNNs). We have tested our model across diverse mass spectra and have observed its superiority over other existing models. MoMS-Net considers substructure at the graph level, which facilitates the incorporation of long-range dependencies while using less memory compared to the graph transformer model.
}


%%\pacs[JEL Classification]{D8, H51}

%%\pacs[MSC Classification]{35A01, 65L10, 65L12, 65L20, 65L70}

\maketitle

\section{Introduction}
Mass spectrometry (MS) \cite{mass_03, mass_13} is an indispensable analytical method for the identification of molecular structures in unknown samples \cite{mass_16, mass_14, mass_18}. In this technique, a molecule undergoes ionization, and its fragment ions are measured by a mass analyzer, which captures information regarding the mass-to-charge ratio (m/z). By analyzing the mass spectrum, which provides the m/z values and their relative intensities, it is possible to infer the molecular structure of the original chemical. 

Modeling the fragmentation patterns for ionized molecules in order to analyze the mass spectrum is challenging. While some domain knowledge-based rules can be useful for certain types of molecules, it becomes difficult to apply them to smaller fragments with diverse functional groups.

The interpretation of mass spectra typically relies on library search, which compare the spectra with a large database of known molecules \cite{mass_95, mass_94}. While there are various extensive mass spectra libraries available, such as the National Institute of Standards and Technology (NIST) \cite{nist_17}, Wiley  \cite{wiley_16}, and Mass Bank of North America (MoNA) \cite{mona2021massbank}, the search-based method is limited by its ability to access known materials and does not provide information on the mass spectra of new molecules. An alternative way is to use $de~ novo$ techniques, which aim to directly predict the molecular structure based on the input spectrum \cite{fp_20, peptide_94, peptide_17}. However, these methods often have low accuracy and are challenging to use effectively \cite{mass_15}.

An approach to address the coverage issue in library search is to enhance existing libraries by incorporating predicted mass spectra generated by a model. Mass spectrum prediction models utilize either quantum mechanical calculations \cite{qm_16,qm_13, qm_12, qm_17}, or machine learning techniques \cite{ml_16}. These methods aim to predict the fragmentation patterns that occur after ionization. Quantum mechanical calculations require extensive computation of electronic state, but they are computationally inefficient. On the other hand, machine learning approaches can provide faster predictions, but they may lack the ability to simulate diverse and detailed fragmentation processes.

Recently, deep learning has been significantly developed in the areas of image recognition and natural language processing. Moreover, there has been a significant surge in interest in applying deep learning to the fields of material science and drug development. Graph Neural Networks (GNNs), in particular, are widely used to predict chemical properties and generate new molecules because molecules, which are comprised of atoms and bonds, can be represented using graph structures, where nodes represent atoms and edges represent bonds.

Several studies have focused on predicting mass spectra using MLP, GNN, and graph transformer\cite{neims_19,cnn_20, gcn_22,young2021massformer, murphy2023efficiently}. J. Wei et al. \cite{neims_19} proposed the NEIMS model, which utilizes fingerprints to map molecules. They employ MLP layers and a bidirectional prediction mode to model fragments and neutral losses in a mass spectrum. B. Zhang et al. \cite{gcn_22} employ a graph convolutional network (GCN) for predicting mass spectra. They initialize the nodes' features using concatenated one-hot vectors representing various atom properties such as atom symbol, degree, valence, formal charge, radical charge, etc. The initial features of edges are also represented using one-hot vectors based on bond type, ring presence, conjugation, and chirality. Multiple GCN layers are applied, and the nodes' representations are pooled to form a graph representation. An MLP layer is then used to predict the mass spectra. A. Young et al. \cite{young2021massformer} proposed the MassFormer framework, which is based on the graph transformer for predicting tandem mass spectra. They employ a graph transformer that calculates pairwise attention between nodes, considers the shortest path distance between two nodes, and incorporates averaged edge information along the shortest path. In the work by M. Murphy et al. \cite{murphy2023efficiently}, they proposed a prediction model that maps an molecular graph to a distribution of probabilities across various molecular formulas using high-resolution mass spectra. This model differs from our task in the sense that high-resolution mass spectra contain additional information about specific peaks that can be used to infer the molecular formulas associated with those peaks.

Motifs, which are important and frequently occurring subgraphs, correspond to functional groups and important fragments in the molecules \cite{motif_02}. Mining motifs can be beneficial in many tasks. There are different approaches to motif mining. One method involves rule-based techniques, where fragmentation rules are defined based on domain knowledge. However, this approach may not cover all possible types of fragments. Alternatively, a motif mining algorithm based on the counting of subgraph structures can be employed. This approach is inspired from Byte Pair Encoding (BPE), a widely used technique in natural language processing (NLP) for tokenizing words into subwords. Motifs can be used to improve the capability for property prediction, drug-gene interaction prediction and molecule generation, due to the strong dependence of a molecule's properties on its molecular structure, particularly the functional groups \cite{hmgnn_22,ngnn_21,gsn_22,hiervae_22,sat_22,cosmig_22}.

\begin{landscape}    

\begin{figure}[ht]%
\centering
\includegraphics[width=1.6\textwidth]{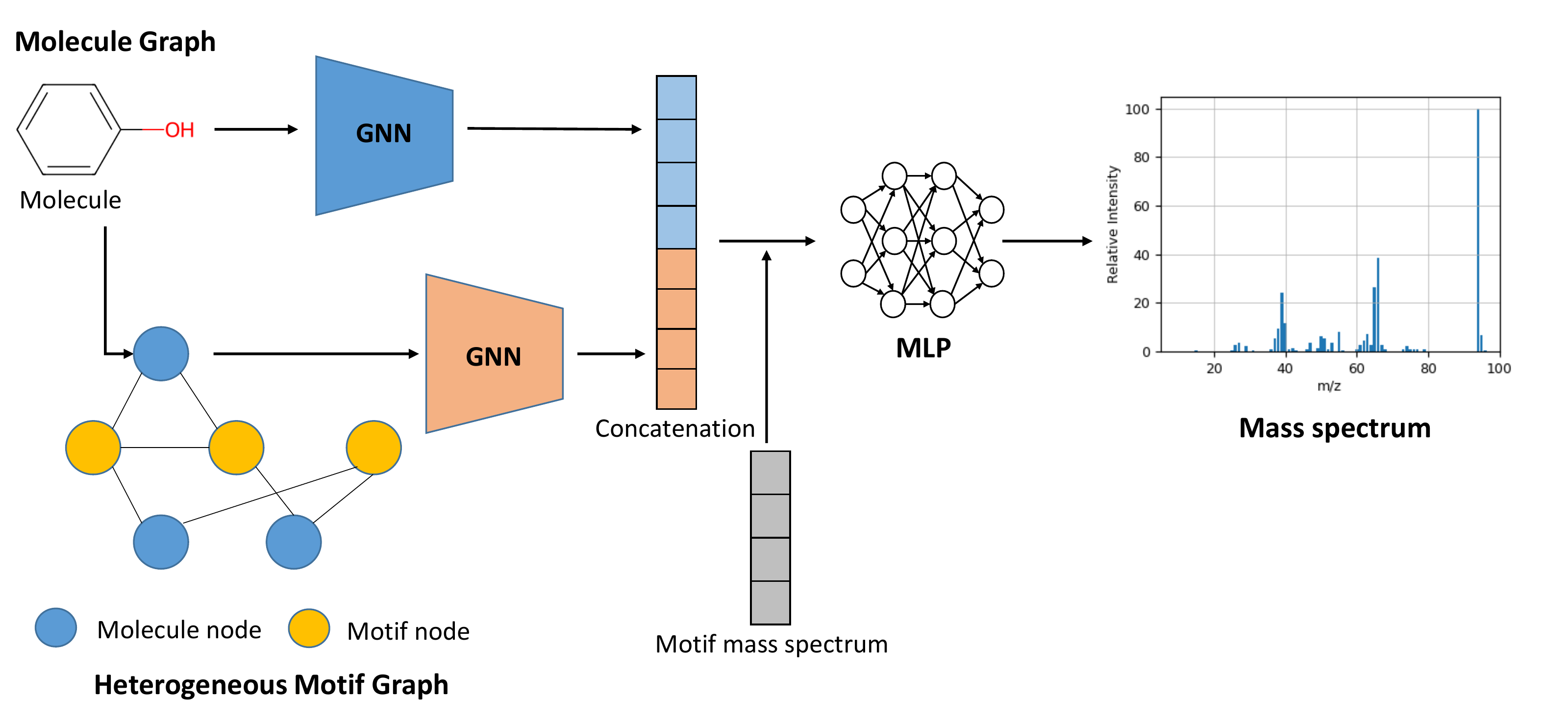}
\caption{Overall architecture of MoMS-Net. The model consists of two GNNs for molecule graph and heterogeneous motif graph. Two graph embeddings from two GNNs are concatenated, and an MLP layer is applied to predict mass spectra.}
\label{hmgnn_intro}
\end{figure}
\end{landscape}

In this work, we propose the Motif-based Mass Spectrum Prediction Network (MoMS-Net) for predicting mass spectra. We utilize motifs in applying GNNs because they are related to the stability of fragment ions and the fragmentation patterns in the mass spectra. The motif vocabulary is constructed by following the merge-and-update method described in Z. Geng et al.\cite{micam_23}. The MoMS-Net model consists of two GNNs, as shown in Fig. \ref{hmgnn_intro}. One is for the molecule graph, which is defined based on the molecule itself. The other is for heterogeneous motif graph, which consists of all molecules in the dataset and the motifs in the motif vocabulary. Before applying the MLP (Multi-Layer Perceptron) layer to predict a mass spectrum, we incorporate the knowledge and characteristics of motifs' mass spectra into our model. GNNs struggle to consider long-range dependencies as node information is updated by pooling neighboring nodes \cite{longrange_18, longrange_20, graphtrans_21}. While deep layers are typically required to incorporate long-range dependencies in GNNs, this can lead to oversmoothing problems where all nodes become similar, resulting in decreased performance \cite{oversmoothing_18, oversmoothing_20, oversmoothing_21}. However, our model can consider the relationship with subgraphs at the graph level, allowing it to effectively incorporate long-range dependency effects. The graph transformer \cite{young2021massformer, graphormer_21} has demonstrated good performance in predicting mass spectra but requires a significant amount of memory during training . In contrast, our model requires less memory than the graph transformer. Ultimately, our model achieves the state-of-the-art performance in predicting mass spectra.
Main contributions of MoMS-Net are as below.
\begin{itemize}
    \item MoMS-Net incorporate relation with substructures in the graph level, so it can consider long-range dependancy effect. In contrast, GNNs cannot consider long-range dependancy as updating node information by pooling neighboring nodes. 
    \item MoMS-Net has the state-of-the-art performance in predicting mass spectra. It shows the highest spectrum similarity compared to other models. 
    \item MoMS-Net requires less memory compared to graph transformer, making it applicable for larger molecules.
    
\end{itemize}

\clearpage

\section{Results}\label{sec2}
\subsection{Overview of the framework}%%\label{subsec1}
The MoMS-Net model consists of two Graph Neural Networks (GNNs) designed to handle both the molecule graph and the heterogeneous motif graph. The molecule graph utilizes a 3-layer Graph Convolutional Network (GCN) \cite{chen2020simple} with RDkit fingerprints \cite{landrum2013rdkit} as input embeddings. On the other hand, the heterogeneous motif graph employs a 3-layer Graph Isomorphic Network (GIN) \cite{xu2018powerful} with input embeddings based on the relationships between molecules and motifs, motifs and motifs, and molecular weights.
The hidden representations obtained from the molecule graph GNN and the heterogeneous motif graph GNN are concatenated to capture the combined information from both graphs. Additionally, the molecular weight distribution of motifs and selected fragments are utilized to further finetune the hidden representations. Finally, an MLP layer is applied to the hidden representation in order to predict the mass spectrum.

\subsection{Spectrum Similarity}
The NIST dataset was divided into three subsets: training (70\%), validation (20\%), and test (10\%), using the Murcko scaffold splitting method \cite{bemis1996properties}. To assess the performance of our model, we made predictions for the mass spectra of the molecules in the test set. We then calculated the spectrum similarity between the target (actual) mass spectra and the predicted mass spectra using the cosine similarity score.
To ensure a fair comparison, we initially normalized both the target and predicted mass spectra. Then, we computed the cosine similarity score between the normalized vectors. Each result has been obtained by conducting the experiments five times, with distinct random seeds for each run. 

The results for the NIST dataset are presented in Table \ref{result}. MoMS-Net demonstrates the best performance compared to other models. Specifically, MassFormer outperforms CNN, WLN and GCN models. Furthermore, we observe that the performance on the FT-HCD dataset is higher compared to the FT-CID dataset. This can be attributed to the larger amount of data available in the FT-HCD dataset. It is commonly known that transformer-based models can achieve better performance when trained on larger dataset\cite{xu2020optimizing}. However, it is noteworthy that MoMS-Net surpasses the performance of MassFormer even in the larger FT-HCD dataset. 

\begin{table}[ht]
\centering
\caption{Cosine Similarity}\label{result}%
    
\begin{tabular}{@{}llll@{}}
\hline
  &FT-CID & FT-HCD \\
\hline
CNN    & $0.356\pm 0.002$   & $0.535\pm 0.002 $   \\
MassFormer     & $0.385\pm 0.005$   & $0.573\pm 0.003 $   \\ %\footnotemark[1] 
WLN    & $0.357\pm 0.001$   & $0.569\pm 0.001 $   \\
GCN    & $0.356\pm 0.001$   & $0.565\pm 0.001 $  \\
MoMS-Net & $\mathbf{0.388\pm 0.002}$   & $\mathbf{0.578\pm 0.001 }$  \\
\hline
\end{tabular}
\footnotetext{Every result is done five times with different random seeds.}
\end{table}

\subsection{Molecule Identification}

To address the coverage issue in spectral library searches, predicting mass spectra is a essential step to augment the existing database. By predicting mass spectra, we can expand the range of compounds and their corresponding spectra available in the spectral library. However, assessing the accuracy of a model in matching predicted spectra with unknown queries is challenging because confirming the identification of the compound requires experimental analysis. To simplify the evaluation process, we can employ a candidate ranking experiment inspired by \cite{neims_19, young2021massformer}. In this experiment, the objective is to accurately associate a query spectrum with the corresponding molecule from a set of candidate spectra. 
The query set comprises authentic spectra from the test set, which are heldout partitions. The reference set consists of spectra collected from distinct origins: predicted spectra in the heldout partition, and real spectra from the training and validation partitions. 
By evaluating the similarity between spectra in the query and reference sets, we calculate a ranking of spectra in the reference set for each query. This ranking, based on the degree of similarity, effectively induces a ranking of candidate structures since each spectrum corresponds to a specific molecule.
Table \ref{top-5} provides a summary of the results obtained from this experiment on the metric, Top-5\%. This metric evaluates whether the true matched candidate is ranked within the top 5\% of all candidates. As the number of candidates per query may vary, the Top-5\% metric is normalized to ensure fair comparison. 
This metric provides insight into the model's ability to accurately identify the correct candidate among a larger set of options.
The results indicate that our model demonstrates comparable performance with MassFormer and higher than other models. This consistent strong performance of our model suggests that it is one of the best performing models in terms of accurately matching query spectra with the correct molecule. Our model can be utilized for augmenting spectral libraries holds promise to address the coverage issue.

\begin{table}[ht]
\centering
\caption{Top-5\% scores on the ranking task}%
\label{top-5}
    
\begin{tabular}{@{}llll@{}}
\hline
  &FT-CID & FT-HCD \\
\hline
CNN    & $0.802\pm 0.008$   & $0.778\pm 0.004 $   \\
MassFormer     & $\mathbf{0.850}\pm 0.016$   & $0.830\pm 0.007 $   \\ 
WLN    & $0.736\pm 0.011$   & $0.812\pm 0.008 $   \\
GCN    & $0.728\pm 0.0.016$   & $0.802\pm 0.008 $  \\
MoMS-Net & ${0.824\pm 0.002}$   & $\mathbf{0.840\pm 0.010 }$  \\
\hline
\end{tabular}
\footnotetext{Every result is done five times with different random seeds.}

\end{table}

\subsection{The Effect of Motif Vocabulary Size}
\begin{figure*}[t]
    \centering
    \subfigure[]{    
        \includegraphics[width=0.4\columnwidth]{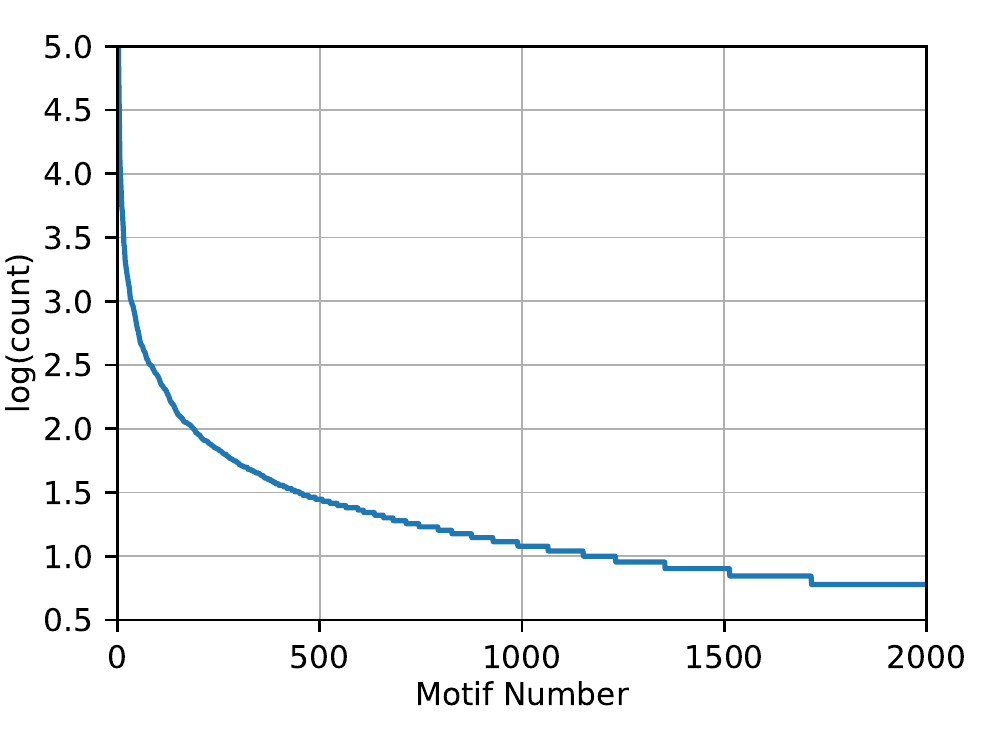}%          
        %\caption{a}
    }
    \subfigure[]{
        \includegraphics[width=0.4\columnwidth]
        {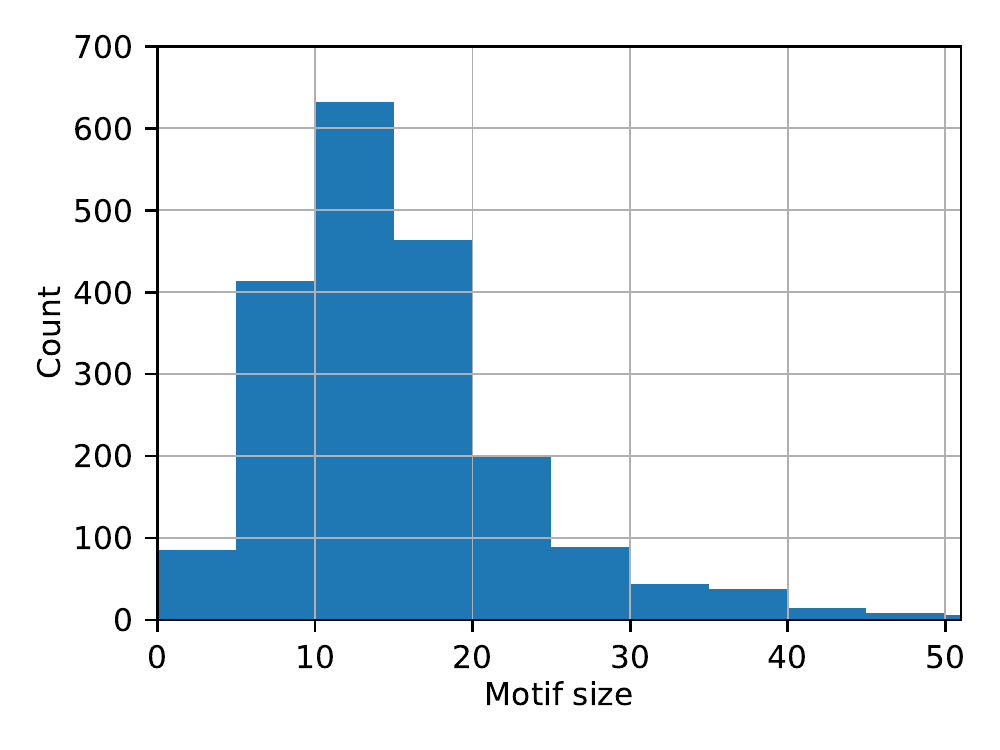}%        
    }

    \subfigure[]{        
        \label{motif_exams}
        \includegraphics[width=0.4\textwidth]{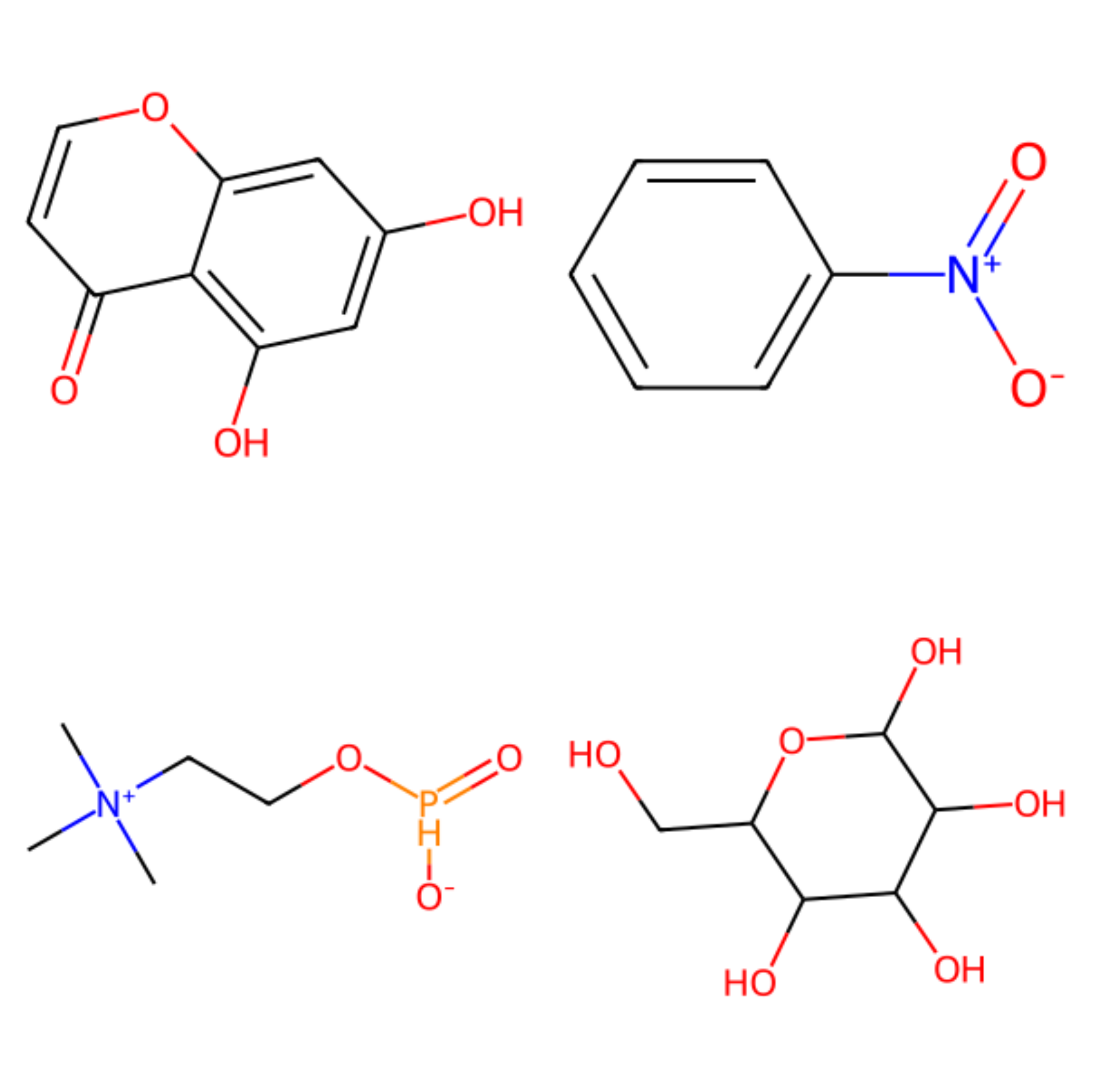} 
    }
    \hspace{2cm}
    \subfigure[]{   
    % \begin{minipage}{1\textwidth}
        \centering
        \includegraphics[width=0.55 \textwidth]{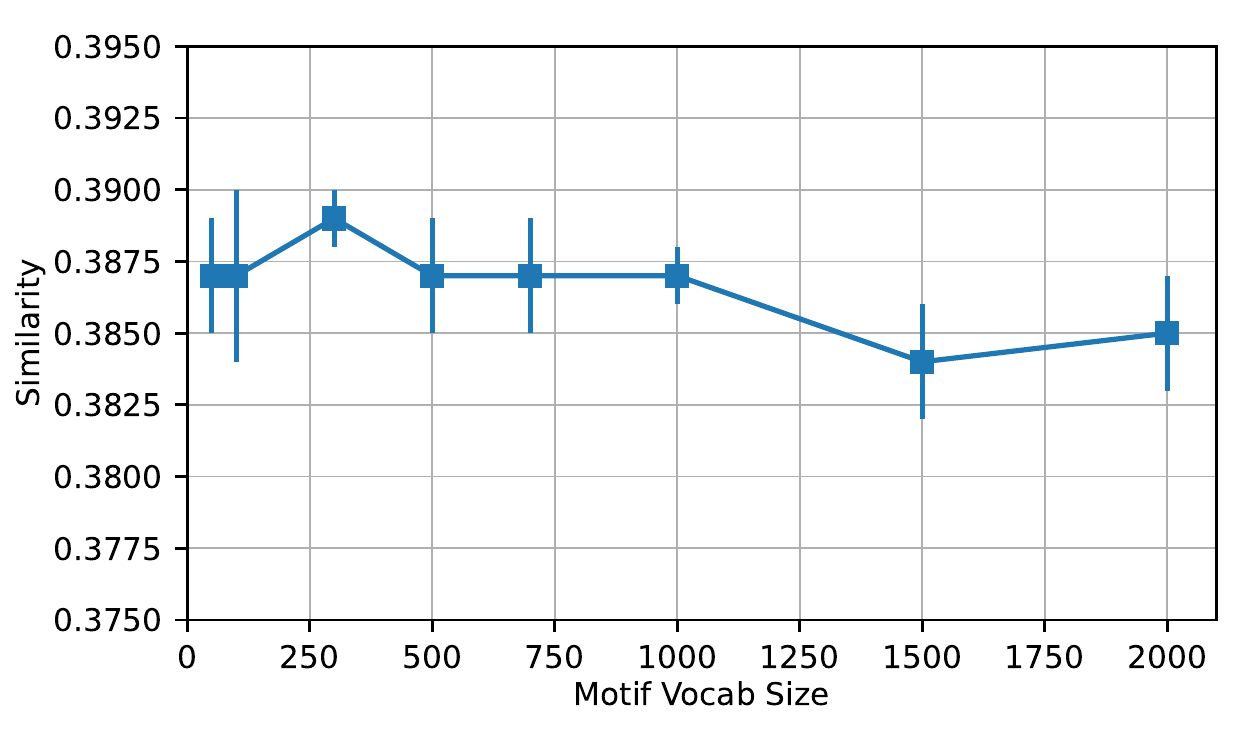} 
    % \end{minipage}
    }
        
    \caption{ (a) Frequency of generated motifs  (b) The distribution of motif size (c) Some examples of large motifs (d) Cosine Similarity according to Motif Size. The frequency of motif is decreased exponentially as motif number and most motif has size of 5 to 20 atoms. Data-driven motif generation method can generate large motifs which have various functional groups. The model achieves its best performance when the motif size is adjusted to 300. However, as the motif size surpasses 1000, the performance starts to decline.}
    \label{motif_freq}
\end{figure*}

We utilized the merge-and-update method to generate motifs from the dataset \cite{micam_23}. The top $K$ most frequent subgraphs were chosen as motifs. We examined the resulting motif vocabulary and observed a decreasing exponential trend in the frequency count as the number of motifs increased, as shown in Fig. \ref{motif_freq}. The most frequent subgraphs are small and stable fragments such as ``$\mathrm{CC}$", ``$\mathrm{CCCC}$",``$\mathrm{CO}$" and benzene ring ($\mathrm{C_6 H_6}$). Our approach allowed for the generation of motifs of various types and sizes, with a higher occurrence of motifs containing 5 to 15 atoms. Fig. \ref{motif_exams} displays several examples of large motifs with distinct structures. We performed tests with different sizes of motif vocabularies and observed that when the motif size exceeded 1,000, the cosine similarity began to decrease. This decline can be attributed to the inclusion of trivial motifs in the heterogeneous motif graph as the motif vocabulary size increased. Hence, in this study, we set the motif vocabulary size to 300.

\subsection{Ablation Study}
We compare GNN architectures for the prediction of mass spectra. As shown in Table \ref{gnn_arch}, we can see that GCN performs better than GIN. In the MoMS-Net model, GIN is utilized for the heterogeneous motif graph, while both GIN and GCN are compared for the molecular graph. Interestingly, when the MoMS-Net model employs GIN instead of GCN for the molecular graph, it exhibits similar performance. 

\section{Discussion}\label{sec3}
The analysis of mass spectra plays a crucial role in identifying molecular structures in material chemistry and drug discovery. Search-based methods are widely employed for mass spectra analysis. However, they often suffer from a coverage issue. To address this problem, it is necessary to generate mass spectra using a model to augment the database. 

\begin{figure}
\centering
\includegraphics[width=1.0 \textwidth ]{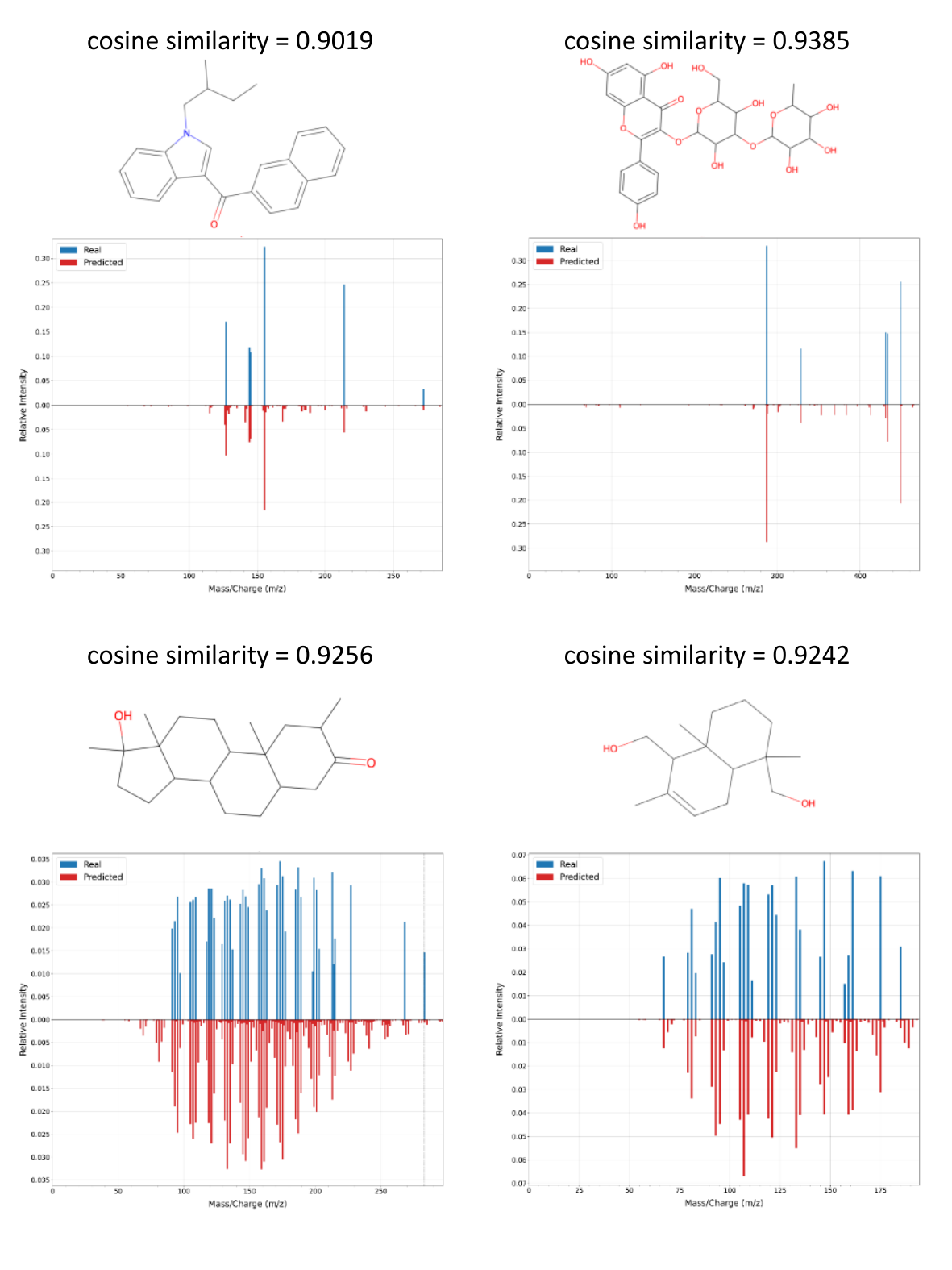}
\caption{True and predicted spectra for four molecules. Predicted spectra have similar patterns for complex molecular structure but have lower intensity because of many smaller peaks. }
\label{mass_spectra}
\end{figure}

MoMS-Net demonstrates the capability to accurately predict mass spectra for complex molecules, as shown in Fig. \ref{mass_spectra}. Molecules containing conjugated aromatic rings are known to be highly stable, resulting in a smaller number of peaks in their mass spectra. On the other hand, molecules without aromatic rings tend to exhibit a greater number of peaks. Our model is effective in predicting both aromatic and nonaromatic compounds accurately. However, it should be noted that there is a limitation in terms of the abundances of the main peaks in the predicted mass spectra. Our model tends to generate more smaller peaks, which can result in a decrease in the intensity of the main peak after normalization.

In Table \ref{param_mem}, the information regarding the number of model parameters and memory allocation is presented. It is observed that all models have similar numbers of model parameters, indicating comparable complexity in terms of the model architecture and parameter count.
However, a notable difference is observed in the memory allocation between MassFormer and MoMS-Net. Despite MassFormer having a smaller batch size, it requires a similar amount of memory allocation compared to MoMS-Net. This suggests that MassFormer consumes a significant amount of memory during its execution.
On the other hand, MoMS-Net demonstrates better performance while utilizing less memory compared to MassFormer. This efficiency in memory usage allows MoMS-Net to handle larger molecules and proteins effectively.

\begin{table}[ht]
\centering
\caption{Number of Parameters and Memory Allocation }\label{param_mem}%
\begin{tabular}{@{}llll@{}}
\hline
  & \# of Parameters & Memory Allocation(MB) & Batch Size \\
\hline
CNN    & 1.46E+07& 717 & 512  \\
MassFormer     & 1.36E+07& 1340& 50  \\
WLN    &1.23E+07 & 1519& 1024  \\
GCN    & 1.31E+0.7& 973& 1024  \\
MoMS-Net & 1.82E+07& 1519& 1024  \\
\hline
\end{tabular}
\end{table}

In this study, we proposed the MoMS-Net model, which incorporates motifs to predict mass spectra from molecular structures. Motifs play a important role in the task of predicting molecular properties as they are directly associated with the functional groups present in the molecule and provide valuable information on the relationships between molecules.
We applied the merge-and-update method to generate a motif vocabulary from the dataset to represent various motif sizes and functional groups. We conducted tests with different sizes of motif vocabularies and varying model architectures. MoMS-Net outperforms other deep learning models in predicting mass spectra from molecular structures. It effectively considers long-range dependencies by incorporating motifs at the graph level even though GNNs have limitations in considering long-range dependencies. Additionally, our model requires less memory compared to the graph transformer. We found that real mass spectra of motifs are useful in predicting the mass spectra of molecules, although the predicted mass spectra may contain more small and false peaks. In future work, we strive to enhance the initialization method of mass spectra for motifs and incorporate regularization techniques to prevent false peaks. Furthermore, we plan to apply MoMS-Net to larger molecules and proteins.

\clearpage

\section{Methods}\label{sec4}
\subsection{Dataset}
We use the NIST 2020 MS/MS dataset for training and evaluation. NIST dataset is a widely used due to its extensive coverage and convenient use in the mass spectrum analysis process. Mass spectra depend on the acquisition conditions. We only use spectra from Fourier Transform (FT) instruments because of the large amount of data available, and we consider their collision cell type (CID or HCD). The information regarding the dataset is summarized in Table \ref{tab_nist}.

\subsection{Generation of Motif Vocabulary}
A motif refers to the most frequent substructure, and some motifs are correspond to functional groups of molecules. To construct a motif vocabulary, we apply the merge-and-update method introduced by A. Young et al. \cite{micam_23} to identify common patterns from a given dataset $D$.
The goal is to learn the top $K$ most frequent subgraphs from dataset $D$, where $K$ is a hyperparameter. Each molecule in $D$ is represented as a graph, $\mathcal{G=(V,E)}$, where atoms and bonds correspond to nodes $(\mathcal{V})$ and edges $(\mathcal{E})$. Initially, we consider each atom from the molecules as a single fragment.\\
We merge two adjacent fragments, $\mathcal{F}_i$ and $\mathcal{F}_j$, to create a new fragment, $\mathcal{F}_{ij} = \mathcal{F}_i \oplus \mathcal{F}_i $, using a defined operation ``$\oplus$". The merging process involves iteratively updating the merging graphs, $\mathcal{G}_M^{(k)}(\mathcal{V}_M^{(k)}, \mathcal{E}_M^{(k)})$ at the $k^{th}$ iteration ($k = 0, \cdots, K-1$). If the most frequent merged fragment, $\mathcal{F}_{ij}$, is valid, it is added to the motif vocabulary $\{\mathcal{M}\}$. This process is iterated for $K$ iterations to construct the motif vocabulary.

\subsection{Construction of the Heterogeneous Motif Graph}
The heterogeneous motif graph is generated by combining molecule nodes from the molecular dataset and motif nodes from the motif vocabulary. This graph consists of two types of edges connecting the nodes.
One type is the molecule-motif edge, which is created when a molecule contains that motif.
The other type is the motif-motif edge, which is established when two motifs share at least one atom.
To differentiate the importance of these edges, different weights are assigned based on their types according to Z. Yu et al. \cite{hmgnn_22}.
For the molecule-motif edge, the weight is computed using the term frequency-inverse document frequency (TF-IDF) value. For the motif-motif edges, the weight is calculated as the co-occurrence information point-wise mutual information (PMI). So the edge weight $A_{ij}$ between two nodes ($i, j$) is represented as

\begin{align} \label{weight2}
    %\begin{split}
        A_{ij}= \left \{ \begin{array} {cc} \text{PMI}_{ij},  &\text{if i, j are motifs} \\ \text{TF-IDF}_{ij}, &\text{if i or j is a motif} \\ 0, &\text{Otherwise} \end{array} \right .
    %\end{split}       
\end{align}

The PMI value is calculated as 
\begin{align}\label{weight_pmi}    
    \begin{split}
        & \text{PMI}_{ij}= \text{log}\frac{p(i,j)}{p(i)p(j)} \\
        & p(i,j) = \frac{N(i,j)}{M}, p(i)=\frac{N(i)}{M}, p(j)=\frac{N(j)}{M},
    \end{split}        
\end{align}

where $N(i,j)$ is the number of molecules that have motif $i$ and motif $j$. $M$ is the total number of molecules, and $N(i)$ is the number of molecules with motif $i$.

\begin{align}\label{weight_tfidf}
    \begin{split}    
        \text{TF-IDF}_{ij} = C(i)_{j}\left(\text{log}\frac{1+M}{1+N(i)} +1 \right),
    \end{split}
\end{align}
where $C(i)_{j}$ is the number of frequency that the motif occurs in the molecule $j$. 

\subsection{Heterogeneous Motif Graph Neural Networks}
We apply two different GNNs for molecule graphs and heterogeneous motif graph. The molecule graph represents each atom and bond as nodes and edges, respectively. We utilize a 3-layer Graph Convolutional Network (GCN) to update the atom-level representations. To encode the atom and bond features, we employ the Deep Graph Library (DGL) package \cite{wang2019deep}, which supports embedding them as either one-hot encoding or numerical values. For the heterogeneous motif graph, we employ the other 3-layer Graph Isomorphism Network (GIN). The total number of nodes in the heterogeneous graph is the sum of the number of molecules ($\vert N \vert$) and the size of the motif vocabulary ($\vert V \vert$). The node feature in the heterogeneous motif graph is represented by the occurrence of motifs and molecule weight for the node. To represent the occurrence of motifs in molecules and other motifs, we create a vector of size $\vert V \vert$, where the values indicate motif occurrences. We apply a linear layer and concatenate it with the molecule weight.\\
A heterogeneous motif consists of all molecule nodes and motif nodes. Since the number of molecules can be large (e.g., 27K for CID and 232K for HCD), computational resource limitations may arise. To tackle this challenge, we employ an edge sampler to decrease the size of the heterogeneous motif graph. We employ a breadth-first algorithm for hop-by-hop sampling from a starting node \cite{hmgnn_22}. We use a 3-hop sampler, denoted as [$s_1,s_2,s_3$], where $s_i$ represents the number of nodes to be sampled. The first-hop neighbors of molecule nodes are motif nodes only. Before applying GINs, we first utilize a 2-layer MLP for input embedding.

\subsection{Mass Spectra of Motif}
After obtaining the graph embeddings for the heterogeneous motif graphs, we incorporate additional information from the mass spectra of motif. This is because the fragmentation patterns in mass spectra are associated with the motif structure. We construct the mass spectra of motifs, taking into account the isotope effect of the molecular ion. Additionally, we incorporate a few fragments generated from RDKit software \cite{landrum2013rdkit} into the motif mass spectra.

\subsection{Objective Function}
Cosine similarity is commonly used in mass spectrum library search to compare and quantify the similarity between mass spectra \cite{mass_94}. So we choose cosine distance as loss function as Eq. \ref{loss}.

\begin{equation}
\label{loss}
    \mathrm{CD}(\mathbf{I},\mathbf{\hat{I}}) = 1 - \frac{\sum_{k=1}^{M_{max}}{I_{k} \cdot {\hat{I_k}}}}{\| \sum_{k=1}^{M_{max}}{I_{k}}^2 \| \cdot \| \sum_{k=1}^{M_{max}}{\hat{I}_{k}}^2 \|}
\end{equation}
where $\mathbf{I}$ and $\hat{\mathbf{I}}$ are vectors of intensities versus m/z for reference and predicted spectrum.

\subsection{Evaluation Metrics}%%\label{subsec1}
The mass spectrum is represented as a vector with a length corresponding to the m/z range, along with intensity values. To measure spectrum similarity, we compute the cosine similarity score between the target and predicted spectra after normalization.
\begin{equation}
    \mathrm{Similarity}(\mathbf{I},\mathbf{\hat{I}}) = \frac{\sum_{k=1}^{M_{max}}{I_{k} \cdot {\hat{I_k}}}}{\| \sum_{k=1}^{M_{max}}{I_{k}}^2 \| \cdot \| \sum_{k=1}^{M_{max}}{\hat{I}_{k}}^2 \|}.
\end{equation}

\section*{Acknowledgments}\label{sec13}

This study was supported by Institute of Information \& communications
Technology Planning \& Evaluation (IITP) grant (No.2021-0-01343, Artificial
Intelligence Graduate School Program in Seoul National University), and
also supported by the National Research Foundation of Korea (NRF) funded
by the Ministry of Education (2022R1A6A3A01087603, 2022R1A3B1077720,
2022M3C1A3081366). All supports were funded by the Korea government(MSIT).

\clearpage

\backmatter

\begin{appendices}

\section{The Dataset}\label{secA1}
We use the NIST 2020 MS/MS dataset. Table \ref{tab_nist} shows the number of spectra and compounds according to their collision cell type (CID or HCD).

\begin{table}[ht]
\caption{NIST MS dataset}\label{tab_nist}%
\begin{tabular}{@{}llll@{}}
\toprule
  Collision Type & \# Spectra & \# Compounds \\
\midrule
FT-CID    & 27,026   & 18,257   \\
FT-HCD    & 322,372  & 19,620   \\
\botrule
\end{tabular}
%\footnotetext{footnote}
\end{table}

\section{GNN architectures}
In the Table \ref{gnn_arch}, MoMS-Net(GIN) uses a GIN for a molecule graph and another GIN for a hetergeneous motif graph. MoMS-Net(GCN) uses a GCN for a molecule graph and a GIN for a heterogeneous motif graph.
\begin{table}[ht]
\centering
\caption{Cosine Similarity according to GNN Architecture }\label{gnn_arch}%
\begin{tabular}{@{}llll@{}}
\hline
  & FT-CID & FT-HCD  \\
\hline
GIN    & $0.352\pm 0.004$ & $0.558\pm 0.001$  \\
GCN     & $0.356\pm 0.002$& $0.565\pm 0.001$  \\
MoMS-Net(GIN)    &$0.389\pm 0.002$ & $0.575\pm 0.001$  \\
MoMS-Net(GCN)    & $0.388\pm 0.002$ & $0.578\pm 0.001$ \\
\hline
\end{tabular}
% \end{center}
\end{table}

%%=============================================%%
%% For submissions to Nature Portfolio Journals %%
%% please use the heading ``Extended Data''.   %%
%%=============================================%%

%%=============================================================%%
%% Sample for another appendix section			       %%
%%=============================================================%%

%% \section{Example of another appendix section}\label{secA2}%
%% Appendices may be used for helpful, supporting or essential material that would otherwise 
%% clutter, break up or be distracting to the text. Appendices can consist of sections, figures, 
%% tables and equations etc.

\end{appendices}

%%===========================================================================================%%
%% If you are submitting to one of the Nature Portfolio journals, using the eJP submission   %%
%% system, please include the references within the manuscript file itself. You may do this  %%
%% by copying the reference list from your .bbl file, paste it into the main manuscript .tex %%
%% file, and delete the associated \verb+\bibliography+ commands.                            %%
%%===========================================================================================%%

\bibliography{sn-bibliography}% common bib file
%% if required, the content of .bbl file can be included here once bbl is generated
%%\input sn-article.bbl

\end{document}